\documentclass[conference]{IEEEtran}

\usepackage{amsmath}
\usepackage{amsfonts}
\usepackage{amssymb}

\usepackage{booktabs}
\usepackage{multirow}

\usepackage{graphicx}
\usepackage{url}
\usepackage[svgnames]{xcolor}

\newcommand{\etal}[1]{#1~\textit{et al.}}
\newcommand{\expnumber}[2]{{#1}\mathrm{e}{#2}}

\begin{document}
\title{Parametric Exponential Linear Unit for Deep Convolutional Neural Networks}

% % % % % % % % % % % %
% % Change this after review.
%
\author{\IEEEauthorblockN{Ludovic Trottier}
\IEEEauthorblockA{Department of Computer Science\\ and Software Engineering\\
Laval University\\
Qu\'ebec, Canada\\
Email: ludovic.trottier.1@ulaval.ca}
\and
\IEEEauthorblockN{Philippe Gigu\`ere}
\IEEEauthorblockA{Department of Computer Science\\ and Software Engineering\\
Laval University\\
Qu\'ebec, Canada\\
Email: philippe.giguere@ift.ulaval.ca}
\and
\IEEEauthorblockN{Brahim Chaib-draa}
\IEEEauthorblockA{Department of Computer Science\\ and Software Engineering\\
Laval University\\
Qu\'ebec, Canada\\
Email: brahim.chaib-draa@ift.ulaval.ca}
}

% % % % % % % % % % % %

\maketitle

\begin{abstract}
Object recognition is an important task for improving the ability of visual systems to perform complex scene understanding. Recently, the Exponential Linear Unit (ELU) has been proposed as a key component for managing bias shift in Convolutional Neural Networks (CNNs), but defines a parameter that must be set by hand. In this paper, we propose learning a parameterization of ELU in order to learn the proper activation shape at each layer in the CNNs. Our results on the MNIST, CIFAR-10/100 and ImageNet datasets using the NiN, Overfeat, All-CNN and ResNet networks indicate that our proposed Parametric ELU (PELU) has better performances than the non-parametric ELU. We have observed as much as a 7.28\% relative error improvement on ImageNet with the NiN network, with only 0.0003\% parameter increase. Our visual examination of the non-linear behaviors adopted by Vgg using PELU shows that the network took advantage of the added flexibility by learning different activations at different layers. 
%\keywords{activation function; deep learning}
\end{abstract}


% \fi
\IEEEpeerreviewmaketitle

\section{Introduction}
\label{sec:intro}

% Why is this a new and important problem?

Recognizing objects using light from the visible spectrum is a essential ability for performing complex scene understanding with a visual system. Vision-based applications, such as face verification, robotic grasping or autonomous driving, require the fundamental skill of object recognition for carrying out their tasks. They must first identify the different elements in their surrounding environment in order to create a high-level representation of the scene. Since scene understanding is performed by analyzing the spatial relations and the taxonomy of the representation, the overall performance of the visual system depends on the capability of recognizing objects. Integrating novel object recognition advances for building fully-automated vision systems is one of the first steps towards general visual perception.

%Performing scene understanding is done by analyzing the spatial relations and the taxonomy of the organization, which can only be achieved with a full comprehension of the objects. Integrating the necessary knowledge in object recognition for building fully automated vision systems is one of the first steps towards general visual perception.

% What are the preliminaries

Over the past few years, Convolutional Neural Networks (CNNs) have become the leading approach in computer vision~\cite{krizhevsky2012imagenet,lecun2015deep,vinyals2015show,jaderberg2015spatial,ren2015faster,hosang2016makes}. Through a series of non-linear transformations, CNNs can process high-dimensional input observations into simple low-dimensional concepts. The key principle in CNNs is that features at each layer are composed of features from the layer below, which creates a hierarchical organization of increasingly abstract concepts. Since levels of organization are often seen in complex biological structures, CNNs are particularly well-adapted for capturing high-level abstractions in real-world observations.

The activation function plays a crucial role for learning representative features. The recently proposed Exponential Linear Unit (ELU) has the interesting property of reducing \textit{bias shift}~\cite{clevert2015fast}. Defined as the change of a neuron's mean value due to weight update, bias shift can lead to oscillations and impede learning when not taken into account~\cite{clevert2015fast}. \etal{Clevert}~\cite{clevert2015fast} have shown that either centering the neuron values with a Batch Normalization layer~\cite{ioffe2015batch} or using activation functions with negative values helps to manage this problem. Defined as identity for positive arguments and $a(\exp(h) - 1)$ for negative ones (where $a=1$ in~\cite{clevert2015fast}), ELU's negative values for negative inputs make the activation function a well-suited candidate for reducing bias shift.

%A recently proposed important activation function is the . Is is defined as identity for positive arguments and $a(\exp(h) - 1)$ for negative ones. The parameter $a$ can be any positive value, but is usually set to $1$. Centering the neuron values can be done with the Batch Normalization (BN) method, while adding negative values can be done with parameterizations such as LReLU or PReLU.

% How does your research bring significant new understanding to the field?
Choosing a proper ELU parameterization can however be relatively cumbersome considering that certain parameterizations are more suitable in some networks than others. The objective of this paper is to alleviate this limitation by learning a parameterization of the activation function, which we refer to as the Parametric ELU (PELU). We contribute in the following ways:
\begin{enumerate}
\item We define parameters controlling different aspects of the function and show how to learn them during back-propagation. Our parameterization preserves differentiability by acting on both the positive and negative parts of the function. It has the same computational complexity as ELU and adds only $2L$ additional parameters, where $L$ is the number of layers.
\item We perform an experimental evaluation on the MNIST, CIFAR-10/100 and ImageNet tasks using the ResNet~\cite{shah2016deep}, Network in Network~\cite{lin2013network}, All-CNN~\cite{springenberg2015striving}, Vgg~\cite{simonyan2014very} and Overfeat~\cite{sermanet2013overfeat} networks. Our results indicates that PELU has better performances than ELU.
\item We evaluate the effect of using Batch Normalization (BN) before our PELU activation, and show that BN increases the error rate of ResNet.
\item We experiment with different PELU parameterizations, and show that the proposed one obtains the best performance among the possible parameterizations.
\item We finally show different PELU non-linear behaviors adopted during training by the VGG network. These results highlight the effects of our parameterization in order to better understand the advantage of the activation.
\end{enumerate}

The rest of the paper is organized as follows. We present related works in Section~\ref{sec:related-work} and described our proposed approach in Section~\ref{sec:pelu}. We detail our experimentations in Section~\ref{sec:experiments} and discuss the results in Section~\ref{sec:discussion}. We conclude the paper in Section~\ref{sec:conclusion}.

%Recently popularized by~\citet{krizhevsky2012imagenet} after winning the 2012 ImageNet competition~\citep{ILSVRC15}, their exceptional ability to capture high level abstractions from observations have led many to consider deeper networks. Since then, several attempts have been proposed to further increase the number of layers, such as the 19-layer Vgg~\citep{simonyan2014very}, the 22-layer GoogleNet~\citep{szegedy2015going}, or the most recent 200-layer ResNet~\citep{he2015deep}. However, building very deep neural networks (DNNs) gives rise to the vanishing and exploding gradients, which impede the learning process~\citep{hochreiter1998vanishing}.

\section{Related Work}
\label{sec:related-work}

% What has been done before and why is it different?
Our proposed PELU activation function is related to other parametric approaches in the literature. Parametric ReLU (PReLU)~\cite{he2015delving} learns a parameterization of the Leaky ReLU (LReLU)~\cite{maasrectifier} activation, defined as $\max\{h,0\} + a \min\{h,0\}$ where $a > 0$. PReLU learns a \textit{leak} parameter $a$ in order to find a proper positive slope for negative inputs. This prevents negative neurons from dying, i.e. neurons that are always equal to zero, which is caused by a null derivative that blocks the back-propagated error signal. Based on the empirical evidence that learning the leak parameter $a$ rather than setting it to a pre-defined value (as done in LReLU) improves performance~\cite{he2015delving}, our goal is further improving the performance of ELU by learning a proper parameterization of the function.

%Defined as $\max\{h,0\}$, the Rectified Linear Unit (ReLU) is one of the most popular activation function~\cite{nair2010rectified}. It has interesting properties, such as low computational complexity, non-contracting first-order derivative and induces sparse activations, which have been shown to improve performance~\cite{krizhevsky2012imagenet}. The main drawback of ReLU is its zero derivative for negative arguments. This blocks the back-propagated error signal from the layer above, which may prevent the network from reactivating dead neurons. To overcome this limitation, Leaky ReLU (LReLU) adds a positive slope $a$ to the negative part of ReLU~\cite{maasrectifier}. Defined as $\max\{h,0\} + a \min\{h,0\}$, where $a > 0$, LReLU has a non-zero derivative for negative arguments. Unlike ReLU, its parameter $a$ allows a small portion of the back-propagated error signal to pass to the layer below. By using a small enough value $a$, the network can still output sparse activations while preserving its ability to reactivate dead neurons. In order to avoid specifying by hand the slope parameter $a$, Parametric ReLU (PReLU) directly learns its value during back-propagation~\cite{he2015delving}. As the training phase progresses, the network can adjust its weights and biases in conjunction with the slopes $a$ of all its PReLU for potentially learning better features. Indeed, \etal{He}~\cite{he2015delving} have empirically shown that learning the slope parameter $a$ gives better performance than manually setting it to a pre-defined value.

The Adaptive Piecewise Linear (APL) unit aims learning a weighted sum of $S$ parametrized Hinge functions~\cite{agostinelli2014learning}. One drawback of APL is that the number of points at which the function is non-differentiable increase linearly with $S$. Differentiable activation functions usually give better parameter updates during back-propagation than activation functions with non-differentiable points~\cite{lecun2015deep}. Moreover, although APL has the flexibility to be either a convex or non-convex function, the rightmost linear function is forced to have unit slope and zero bias. This may be an inappropriate constraint which could affect the CNN ability to learn representative features.

Another activation function is Maxout, which outputs the maximum over $K$ affine functions for each input neuron~\cite{goodfellow2013maxout}. The main drawback of Maxout is that it multiplies by $K$ the amount of weights to be learned in each layer. In the context of CNNs where the max operator is applied over the feature maps of each $K$ convolutional layers, the increased computational burden can be too demanding for deep network. Unlike Maxout, our PELU adds only $2 L$ parameters, where $L$ is the number of layers, which makes our activation as computationally demanding as the original ELU function. 

The S-Shaped ReLU (SReLU) imitates the Webner-Fechner law and the Stevens law by learning a combination of three linear functions~\cite{jin2015deep}. Although this parametric function can be either convex or non-convex, SReLU has two points at which it is non-differentiable. Unlike SReLU, our PELU is fully differentiable, since our parameterization acts on both the positive and negative sides of the function. This in turn improves the back-propagation weight and bias updates.

\section{Parametric Exponential Linear Unit}
\label{sec:pelu}

The standard Exponential Linear Unit (ELU) is defined as identity for positive arguments and $a(\exp(h) - 1)$ for negative arguments ($h<0$)~\cite{clevert2015fast}. Although the parameter $a$ can be any positive value, \etal{Clevert}~\cite{clevert2015fast} proposed using $a=1$ to have a fully differentiable function. For other values $a \neq 1$, the function is non-differentiable at $h=0$. Directly learning parameter $a$ would break differentiability at $h=0$, which could impede back-propagation~\cite{lecun2015deep}. 

\begin{figure}[tp]
\centering

\begin{minipage}{\linewidth}
\centering
\begin{minipage}{0.45\linewidth}
\includegraphics[width=\linewidth]{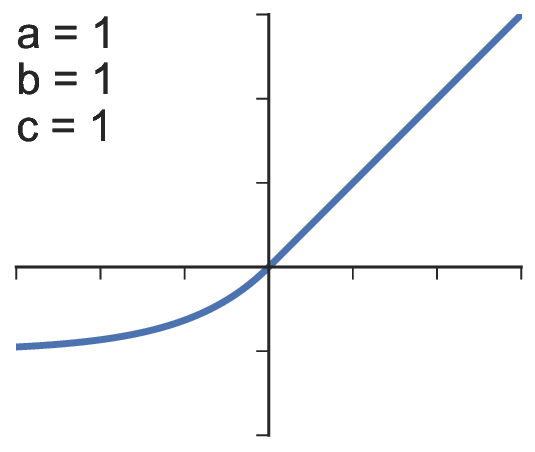}
\end{minipage}
\end{minipage}

\begin{minipage}{\linewidth}
\centering
\begin{minipage}{0.45\linewidth}
\includegraphics[width=\linewidth]{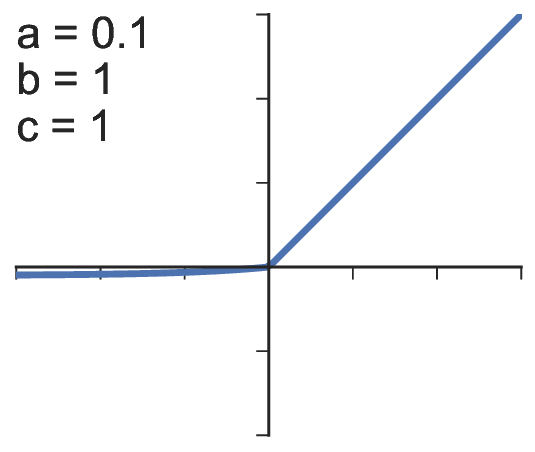}
\end{minipage}
\begin{minipage}{0.45\linewidth}
\includegraphics[width=\linewidth]{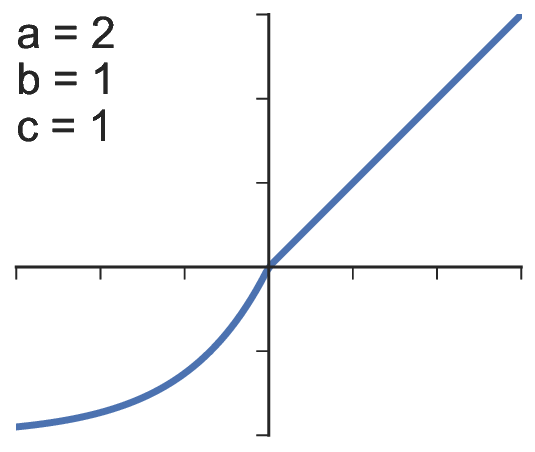}
\end{minipage}
\end{minipage}

\begin{minipage}{\linewidth}
\centering
%Effect of $a$

%\rule{0.8\linewidth}{0.7pt}
\end{minipage}

\begin{minipage}{\linewidth}
\centering
\begin{minipage}{0.45\linewidth}
\includegraphics[width=\linewidth]{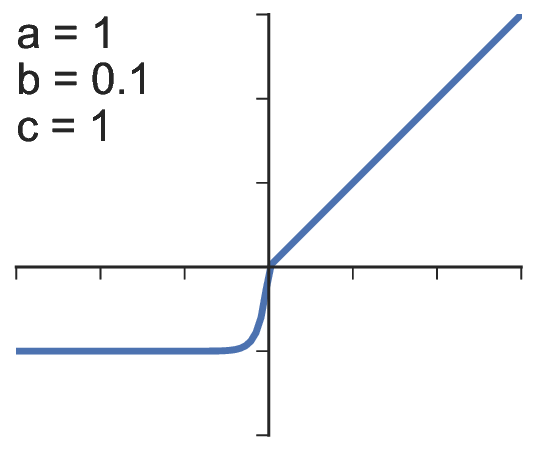}
\end{minipage}
\begin{minipage}{0.45\linewidth}
\includegraphics[width=\linewidth]{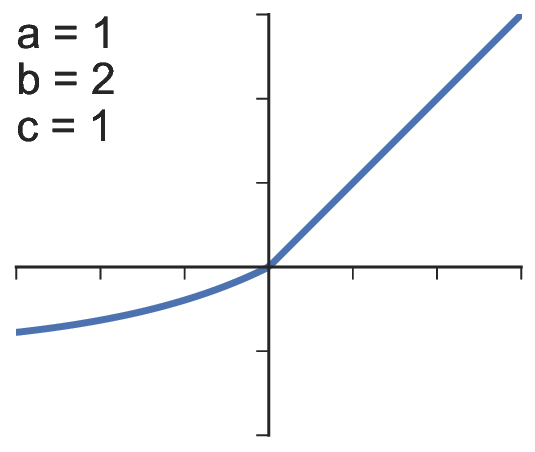}
\end{minipage}
\end{minipage}

\begin{minipage}{\linewidth}
\centering
%Effect of $b$

%\rule{0.8\linewidth}{0.7pt}
\end{minipage}

\begin{minipage}{\linewidth}
\centering
\begin{minipage}{0.45 \linewidth}
\includegraphics[width=\linewidth]{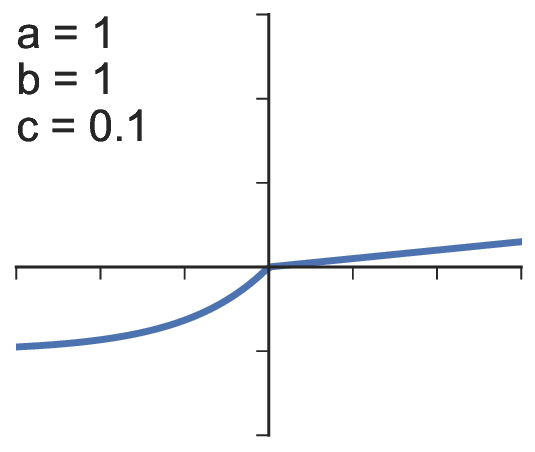}
\end{minipage}
\begin{minipage}{0.45 \linewidth}
\includegraphics[width=\linewidth]{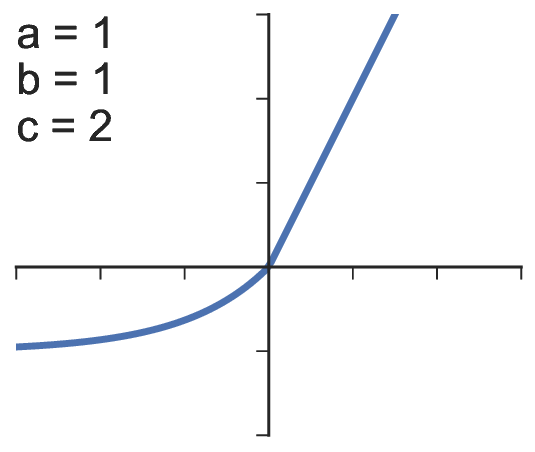}
\end{minipage}
\end{minipage}

\begin{minipage}{\linewidth}
\centering
%Effect of $c$

%\rule{0.8\linewidth}{0.7pt}
\end{minipage}

\caption{Effects of parameters $a$, $b$ and $c$ on the Exponential Linear Unit (ELU) activation function. The original ELU is shown at the top, where $a = b = c= 1$. We show the effect of $a$ on the second row, the effect of $b$ on the third row, and the effect of $c$ on the fourth row. The saturation point decreases when $a$ increases, the function saturates faster when $b$ decreases, and the slope of the linear part increases when $c$ increases.}
\label{fig:effects-of-parameters}
\end{figure}

For this reason, we first start by adding two additional parameters to ELU:
% as follows:
\begin{align}
\label{eq:elu-param1}
f(h) = \begin{cases}
c h & \text{if }  h \geq 0 \\
a (\exp(\frac{h}{b}) - 1) & \text{if } h < 0
\end{cases}
, \quad a, b, c > 0 \, ,
\end{align}
We have $c h $ for positive arguments ($h \geq 0$) and $a (\exp(\frac{h}{b}) - 1)$ for negative arguments ($h<0$). The original ELU can be recovered when $a=b=c=1$. As shown in Figure~\ref{fig:effects-of-parameters}, each parameter controls different aspects of the activation. Parameter $c$ changes the slope of the linear function in the positive quadrant (the larger $c$, the steeper the slope), parameter $b$ affects the scale of the exponential decay (the larger $b$, the smaller the decay), while $a$ acts on the saturation point in the negative quadrant (the larger $a$, the lower the saturation point). Constraining the parameters in the positive quadrant forces the activation to be a monotonic function, such that reducing the weight magnitude during training always lowers the neuron contribution.

Using this parameterization, the network can control its non-linear behavior throughout the course of the training phase. It may increase the slope with $c$, the decay with $b$ or lower the saturation point with $a$. However, a standard gradient update on parameters $a,b,c$ would make the function non-differentiable at $h=0$ and impair back-propagation. Instead of relying on a projection operator to restore differentiability after each update, we constrain our parameterization to always have differentiability at $h=0$. By equaling the derivatives on both sides of zero, solving for $c$ gives 	$c = \frac{a}{b}$ as solution. The proposed Parametric ELU (PELU) is then as follows:
\begin{align}
\label{eq:pelu}
f(h) = \begin{cases}
\frac{a}{b} h & \text{if }  h \geq 0 \\
a (\exp( \frac{h}{b}) - 1) & \text{if } h < 0
\end{cases}
, \quad
a, b > 0
\end{align}
With this parameterization, in addition to changing the saturation point and exponential decay respectively, both $a$ and $b$ adjust the slope of the linear function in the positive part to ensure differentiability at $h=0$.

PELU is trained simultaneously with all the network parameters during back-propagation. Using the chain rule of derivation, the gradients of $f$ with respect to $a,b$ is given by:
\begin{align}
\frac{\partial f(h) }{ \partial a} &= 
\begin{cases}
\frac{h}{b} & \text{if }  h \geq 0 \\
\exp(h/b) - 1 & \text{if } h < 0
\end{cases}
\\ 
\frac{\partial f(h) }{ \partial b} &= 
\begin{cases}
-\frac{ah}{b^2} & \text{if }  h \geq 0 \\
-\frac{a}{b^2}\exp(h/b)  & \text{if } h < 0
\end{cases}
\, .
\end{align}
%
%the derivative of objective $E$ with respect to $a$ and $b$ for one layer is:
%\begin{align}
%\frac{\partial E }{\partial a} &= \sum_{i} \frac{\partial E}{\partial f(h_i)} \frac{\partial f(h_i)}{\partial a}
%, &
%\frac{\partial E }{\partial b} &= \sum_{i} \frac{\partial E}{\partial f(h_i)} \frac{\partial f(h_i)}{\partial b}
%\, ,
%\end{align}
%where $i$ sums over all elements of the tensor on which $f$ is applied. The terms $\frac{\partial E}{\partial f(h_i)}$ are the gradients propagated from the above layers, while $\frac{\partial f(h) }{ \partial a}$ and $\frac{\partial f(h) }{ \partial b}$ are the gradients of $f$ with respect to $a,b$:
%\begin{align}
%\frac{\partial f(h) }{ \partial a} &= 
%\begin{cases}
%\frac{h}{b} & \text{if }  h \geq 0 \\
%\exp(h/b) - 1 & \text{if } h < 0
%\end{cases}
%, & 
%\frac{\partial f(h) }{ \partial b} &= 
%\begin{cases}
%-\frac{ah}{b^2} & \text{if }  h \geq 0 \\
%-\frac{a}{b^2}\exp(h/b)  & \text{if } h < 0
%\end{cases}
%\, .
%\end{align}
For preserving parameter positivity after the updates, we constrain them to always be greater than $0.1$. 

\section{Experimentations}
\label{sec:experiments}

In this section, we present our experiments in supervised learning on the CIFAR-10/100 and ImageNet tasks. 
%Our goal is to show that, with the same network architecture, parameterizing ELU improves the performance. We also provide results with the ReLU activation function for reference. 

\subsection{MNIST Auto-Encoder}

\begin{figure}[t]
\centering
%\begin{minipage}{0.95\linewidth}
%\includegraphics[width=\linewidth]{figures/mnist_aa_train}
%\end{minipage}
\begin{minipage}{0.95\linewidth}
\includegraphics[width=\linewidth]{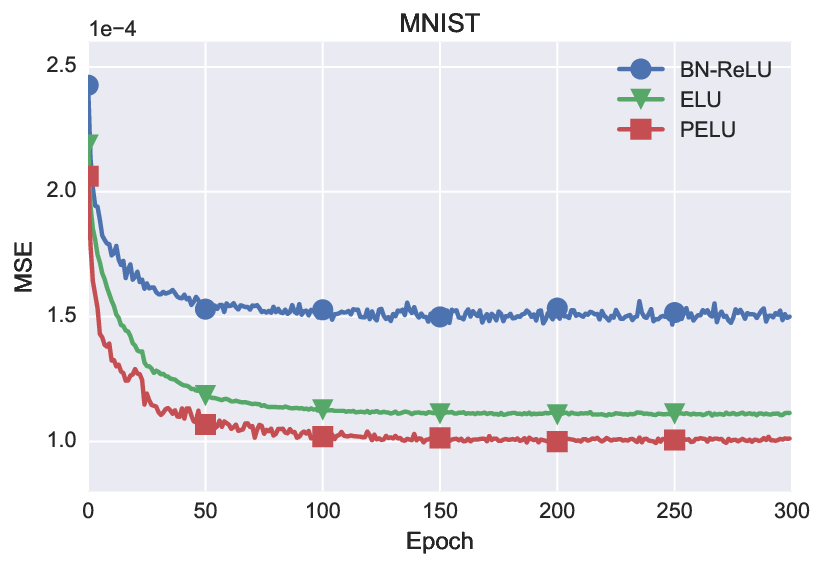}
\end{minipage}
\caption{Auto-encoder results on the MNIST task. We compare PELU to ELU, and include BN-ReLU as additional reference. Compared to ELU, PELU obtained a lower test mean squared error.}
\label{fig:mnist-results}
\end{figure}

As first experiment, we performed unsupervised learning, which is the task of learning feature representations from unlabeled observations. Unsupervised learning can be useful in cases like deep learning data fusion~\cite{srivastava2012multimodal}. For evaluating our proposed PELU activation, we trained a deep auto-encoder on unlabeled MNIST images~\cite{mnistlecun}. We refer to this network as DAA-net. The encoder has four fully connected layers of sizes 1000, 500, 250, 30, and the decoder is symmetrical to the encoder (the weights are not tied). We used Dropout with probability 0.2 after each activation~\cite{srivastava2014dropout}. For ReLU, we put a Batch Normalization (BN) layer before the activation. We trained DAA-Net with RMSProp~\cite{rmsprop} at a learning rate of 0.001, smoothing constant of 0.9 and a batch size of 128.

Figure~\ref{fig:mnist-results} presents the progression of test mean squared error averaged over five tries of DAA-Net on MNIST dataset. These results show that PELU outperformed ELU and ReLU for both convergence speed and reconstruction error. PELU converged approximatively at epoch 75 with a MSE of $\expnumber{1.04}{-4}$, while ELU converged at epoch 100 with a MSE of $\expnumber{1.12}{-4}$ and ReLU at epoch 100 with a MSE of $\expnumber{1.49}{-4}$.

\subsection{CIFAR-10/100 Object Recognition}
\label{ssec:cifar-experiment}

% In a recent study,~\citet{srivastava2015training} noticed that adding more layers to suitably pre-trained networks led to higher training error, which was surprisingly not a consequence of \textit{over-fitting}. The performance degradation observed as depth increases was caused by the network's inability to disable the newly added layers, which would have been possible by learning identity mappings. This lack of capability is non-intuitive since efficient networks can be artificially deepened by stacking identity mappings, without any loss of performance. The existence of such a constructed solution has led~\citet{he2015deep} to consider skipping connections to help disable unnecessary layers, which they refer to as residual blocks. Originally trained with a combination of BN + ReLU, \citet{he2015deep}'s ResNet has recently been extended to ELU activation~\citep{shah2016deep}, with great success.

\begin{figure}
\centering
\includegraphics[width=\linewidth]{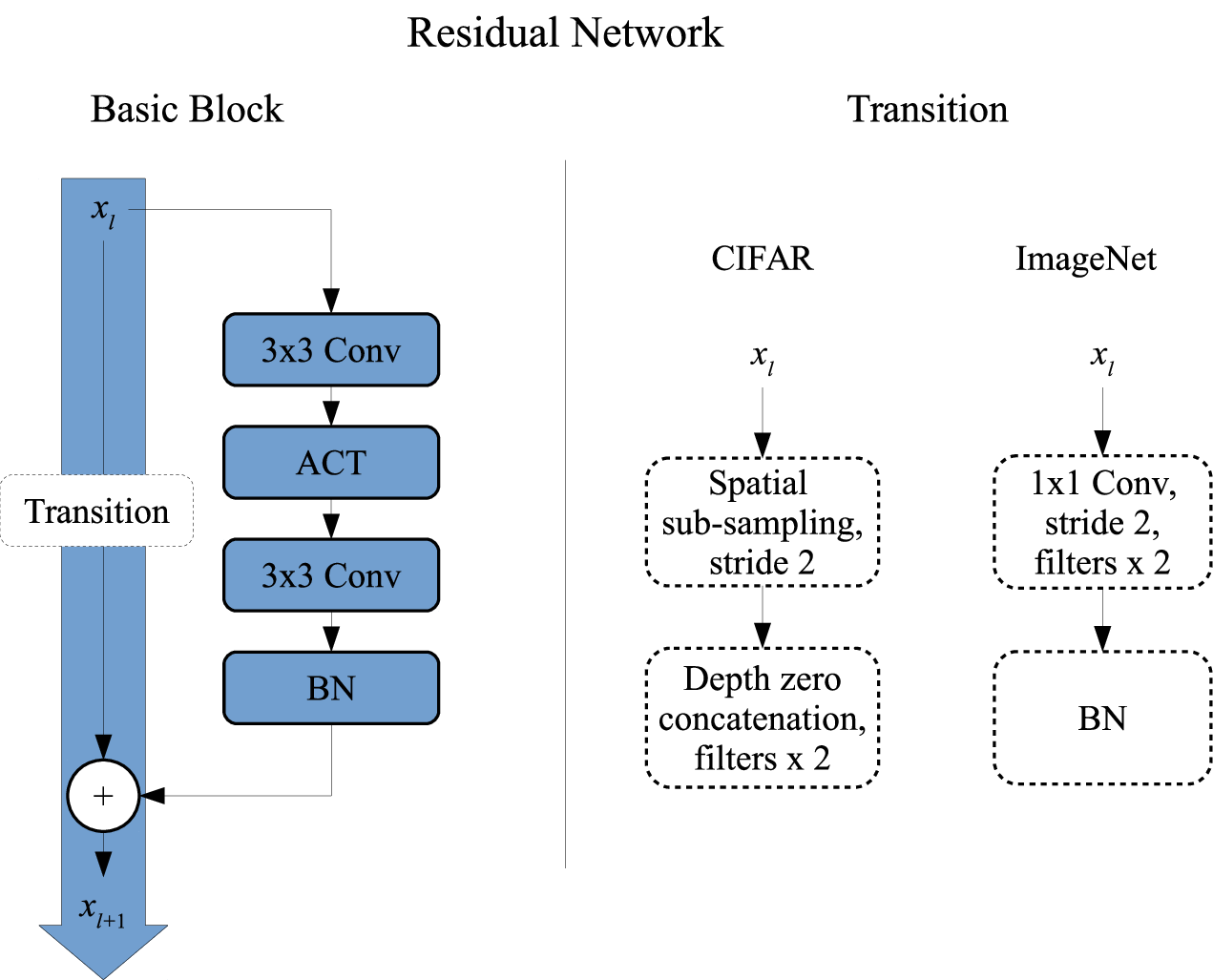}
\caption{Residual network building block structure. On the left, the main basic block structure, and on the right, the transition block structure for reducing the input spatial dimensions and increasing the number of filters. For our CIFAR experiments, we opted for sub-sampling followed by zero concatenation as transition block, while for our ImageNet experiments, we opted for strided convolution followed by batch normalization as transition block.}
\label{fig:resnet-structure}
\end{figure}

\begin{figure}[t]
\centering
\begin{minipage}{\linewidth}
\centering
\begin{minipage}{\linewidth}
\includegraphics[width=\linewidth]{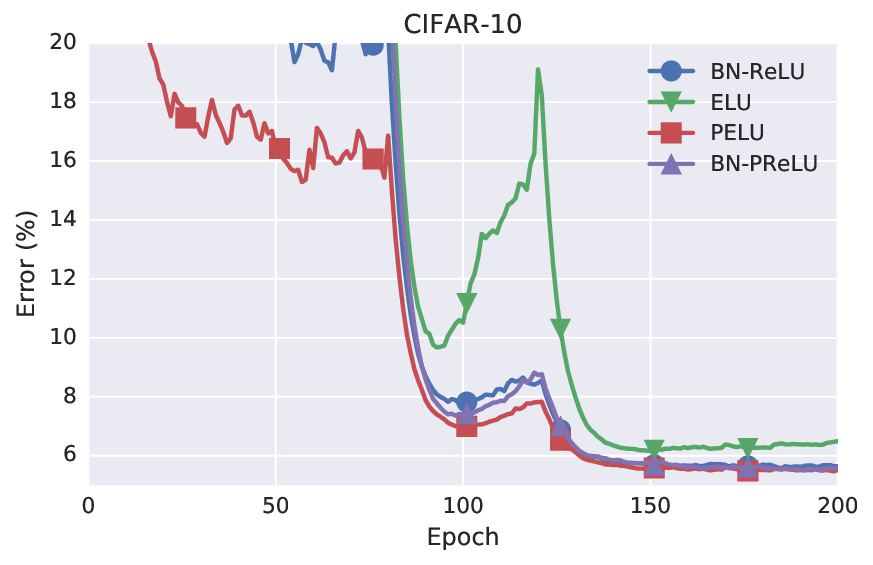}
\end{minipage}
\begin{minipage}{\linewidth}
\includegraphics[width=\linewidth]{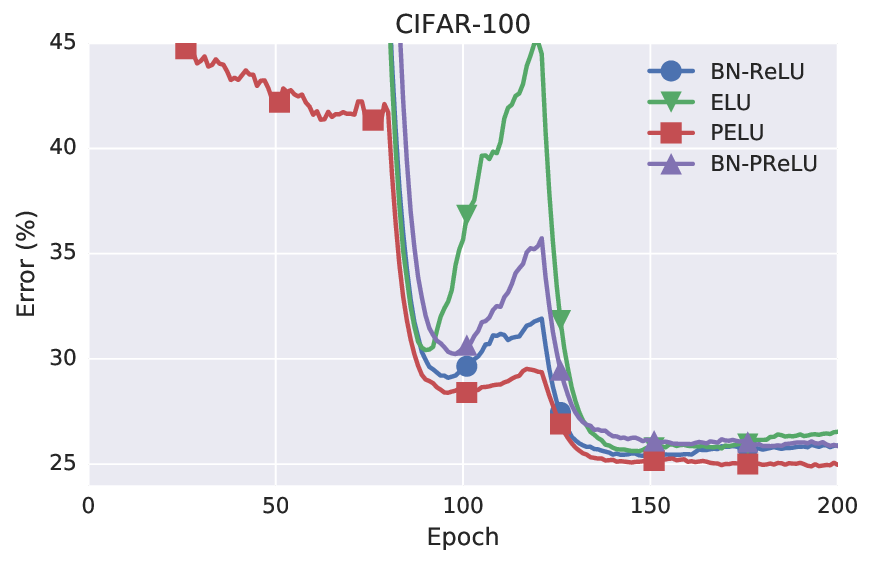}
\end{minipage}
\end{minipage}
\caption{ResNet 110 layers test error (in \%) medians over five tries on both CIFAR-10 and CIFAR-100 datasets. We compare PELU to ELU, and also include BN-ReLU and BN-PReLU as additional references. PELU has a better convergence and lower recognition errors than ELU.}
\label{fig:results-resnet-cifar}
\end{figure}

\begin{table}[t]
\centering
\caption{ResNet 110 layers test error (in \%) on both CIFAR-10 and CIFAR-100 datasets. We report the mean error over the last five epochs and the minimum error over all epochs (inside parenthesis) of the median error over five tries. Our ELU parameterization improves performance, but using BN before the activation worsen the performance.}
%\vspace{4pt}
\begin{tabular}{ccc}
\toprule
ACT & CIFAR-10 & CIFAR-100 \\
\midrule
BN-ReLU & 5.67 (5.41) & 25.92 (24.99)\\
ELU & 6.55 (5.99) & 26.59 (25.08)\\
PELU & \textbf{5.51 (5.36)} & \textbf{25.02 (24.55)} \\
%ReLU & 5.63 (5.38) & 25.47 (24.84) \\
BN-PReLU & 5.61 (\textbf{5.36}) & 25.83 (25.50) \\
\midrule
BN-PELU & 6.24 (5.85) & 26.04 (25.38) \\
BN-ELU & 11.20 (10.39) & 35.51 (34.75) \\
\bottomrule
\end{tabular}
\label{tab:cifar-experiments}
\end{table}

We performed object classification on the CIFAR-10 and CIFAR-100 datasets (60,000 32x32 colored images, 10 and 100 classes respectively)~\cite{krizhevsky2012imagenet}. We trained a 110-layer residual network (ResNet) following Facebook's Torch implementation \verb|fb.resnet.torch|\footnote{ \url{https://github.com/facebook/fb.resnet.torch}}. In order not to favor PELU to the detriment of ELU and BN+ReLU, we performed minimal changes by only replacing the activation function. 

The building block structure for the network is shown in Figure~\ref{fig:resnet-structure}. We show the basic block structure on the left of Figure~\ref{fig:resnet-structure} and the transition block structure on the right of Figure~\ref{fig:resnet-structure}. The ACT module can be PELU, ELU, ReLU or PReLU, with or without BN. The network contains mainly basic blocks, and a few transition blocks for reducing the spatial dimensions of the input image and increasing the number of filters. The ResNet for our CIFAR experiments has a transition block structure with spatial sub-sampling and zero concatenation, while the ResNet for our ImageNet experiments (see Section~\ref{ssec:experiment-imagenet}) has a transition block structure with a strided convolution followed by Batch Normalization.

To train the network, we used stochastic gradient descent with a weight decay of $\expnumber{1}{-3}$, momentum of 0.9 and mini batch-size of 256. The learning rate starts at 0.1 and is divided by 10 after epoch 81, and by 10 again after epoch 122. We performed standard center crop + horizontal flip for data augmentation: four pixels were added on each side of the image, and a random 32 x 32 crop was extracted, which was randomly flipped horizontally. Only color-normalized 32 x 32 images were used during the test phase.

Figure~\ref{fig:results-resnet-cifar} presents ResNet test error (in \%) medians over five tries on both CIFAR datasets. ResNet obtained a minimum median error rate on CIFAR-10 of 5.41\% with BN+ReLU, 5.99\% with ELU, 5.36\% with PELU and 5.26\% with BN-PReLU, while ResNet obtained a minimum median error rate on CIFAR-100 of 24.99\% with BN+ReLU, 25.08\% with ELU, 24.55\% with PELU and 25.50\% with BN+PReLU. In comparison to ELU, PELU obtained a relative improvement of 10.52\% and 2.11\% on CIFAR-10 and CIFAR-100 respectively. It is interesting to note that PELU only adds 112 additional parameters, a negligible increase of 0.006\% over the total number of parameters.

We observed that PELU has a better convergence behavior than ELU. As shown in Figure~\ref{fig:results-resnet-cifar}, ELU has a large test error rate increase at the end of the second stage of the training phase on both CIFAR-10 and CIFAR-100 datasets. Although PELU has also a test error rate increase at the end of the second stage, it does not increase as high as ELU. We further observe a small test error rate increase at the end of the training phase for ELU, while PELU converges in a steady way without a test error rate increase. These results show that training a ResNet with our parameterization can improve the performance and the convergence behavior over a ResNet with ELU activation.

Compared to ReLU, PReLU obtained a smaller minimum median error rate on CIFAR-10 and a smaller average median error rate on CIFAR-100. As shown in Table~\ref{tab:cifar-experiments}, PReLU obtained a minimum median error rate of 5.36 compared to 5.41 on CIFAR-10, and an average median error rate of 25.83 compared to 25.92 on CIFAR-100. Although PReLU obtained the same minimum median error rate than PELU on CIFAR-10, it is significantly higher on CIFAR-100. Note that our main contribution is showing performance improvement over ELU, and that we only add PReLU as an additional reference. Nonetheless, we observe that our PELU parameterization of ELU obtains higher relative improvements than the PReLU parameterization of ReLU.

\subsection{Understanding the effect of Batch Normalization}
\label{ssec:effect-of-bn}

\begin{figure*}[t]
\centering
%\begin{minipage}{0.48\linewidth}
%\centering
%\includegraphics[width=\linewidth]{figures/effect-of-bn_relu_dataset-cifar10-depth-110}
%\end{minipage}
%\begin{minipage}{0.48\linewidth}
%\centering
%\includegraphics[width=\linewidth]{figures/effect-of-bn_relu_dataset-cifar100-depth-110}
%\end{minipage}

\begin{minipage}{0.48\linewidth}
\centering
\includegraphics[width=\linewidth]{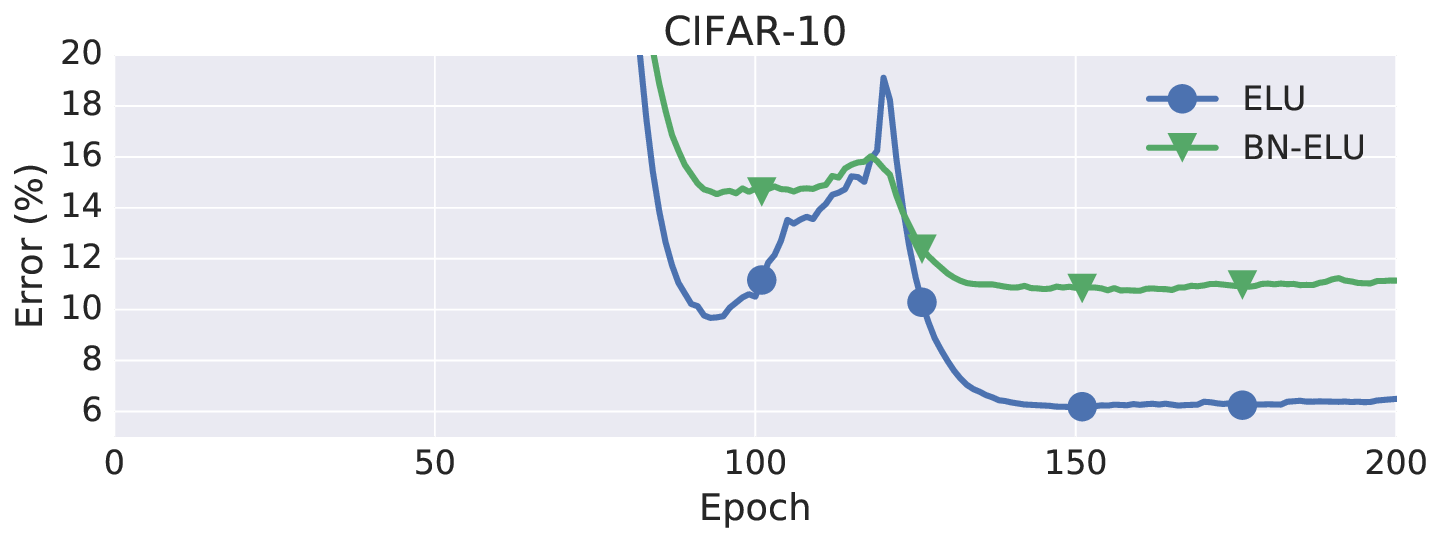}
\end{minipage}
\begin{minipage}{0.48\linewidth}
\centering
\includegraphics[width=\linewidth]{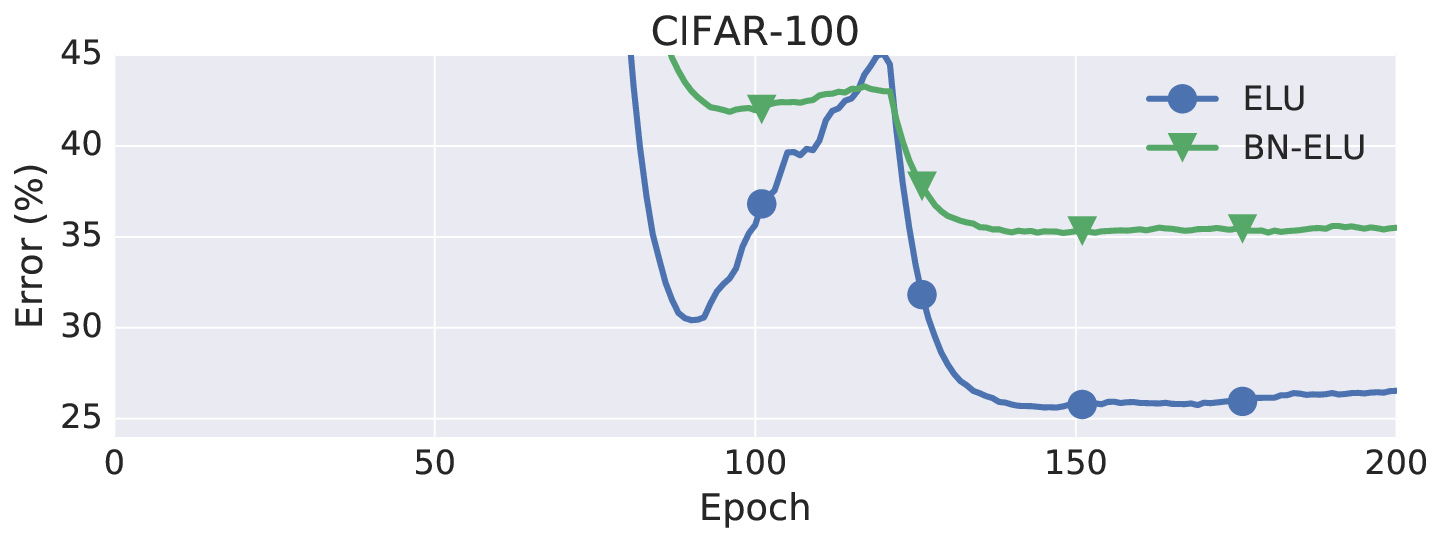}
\end{minipage}

\begin{minipage}{0.48\linewidth}
\centering
\includegraphics[width=\linewidth]{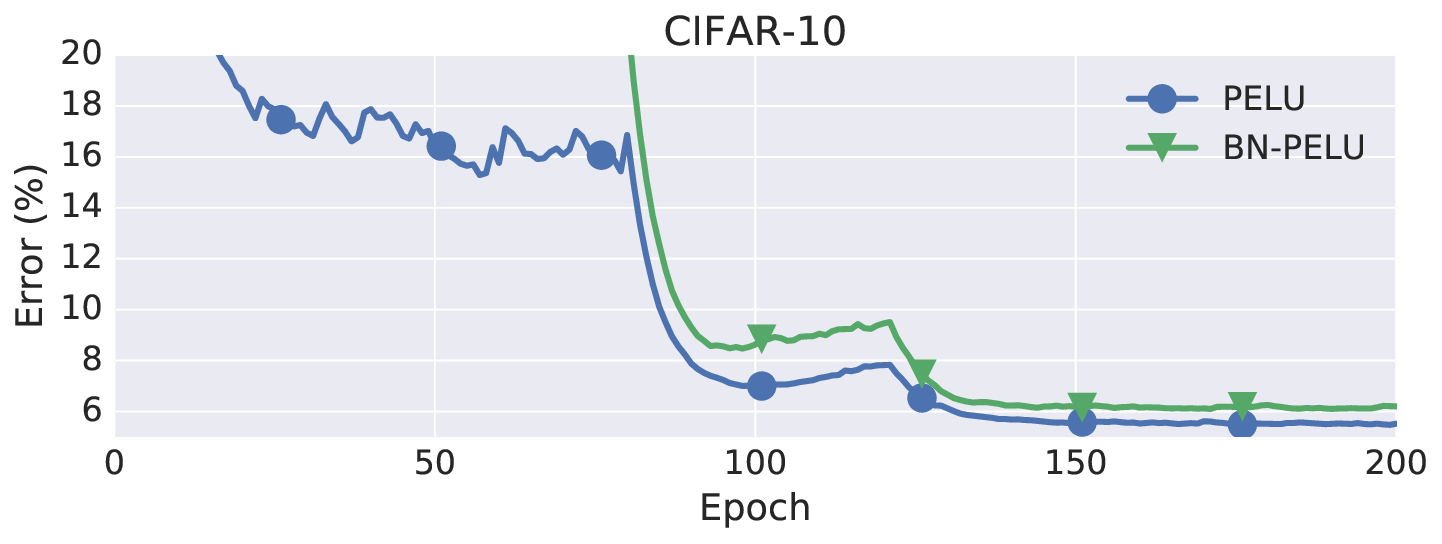}
\end{minipage}
\begin{minipage}{0.48\linewidth}
\centering
\includegraphics[width=\linewidth]{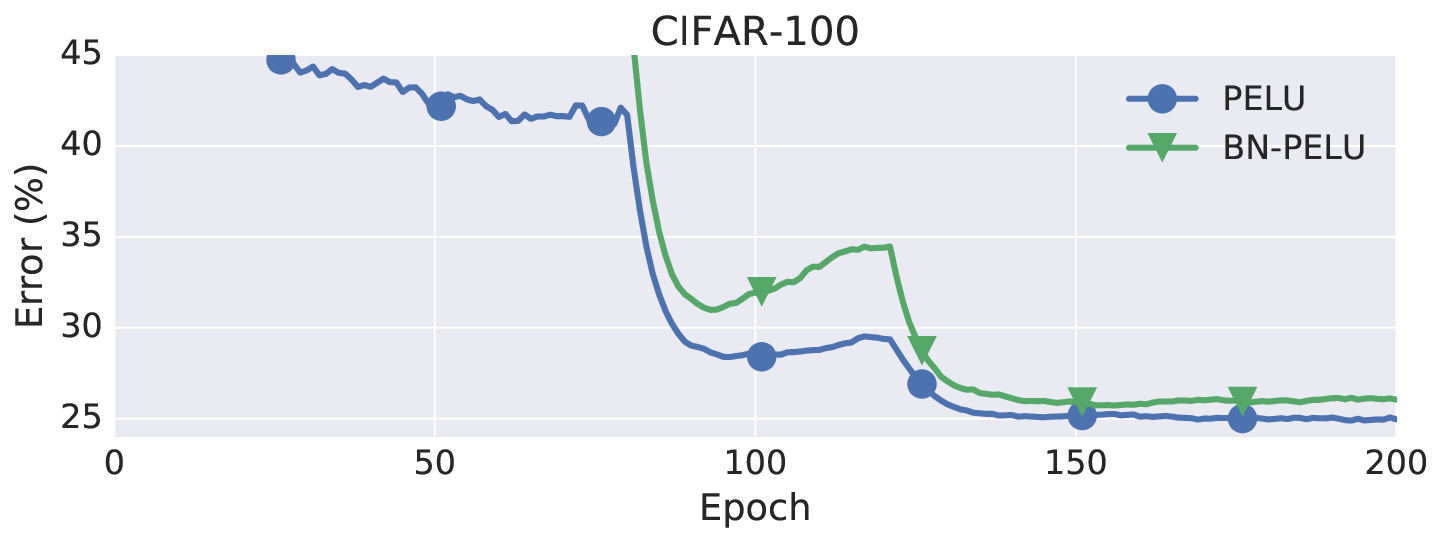}
\end{minipage}

\caption{Effect of using BN before ELU (first row) and PELU (second row) activations in a ResNet with 110 layers on both CIFAR-10 and CIFAR-100 datasets. We show the convergence behavior of the median test error over five tries. In both cases, BN worsen performance of ELU and PELU. Note that we still use BN after the second conv layer, as seen in Figure~\ref{fig:resnet-structure}.}
\label{fig:effect-of-bn}
\end{figure*}

In this section, we show that using BN before our PELU activation has a detrimental effect on its performance. Figure~\ref{fig:effect-of-bn} presents the influence of using BN before ELU and PELU in a ResNet with 110 layers on both CIFAR-10 and CIFAR-100 datasets. We trained the networks using the same framework as in Section~\ref{ssec:cifar-experiment}, but added BN before each activate. Note that in all cases, we use BN after the second convolutional layer in the basic block (see Figure~\ref{fig:resnet-structure}). 

The results show a large error rate increase on both CIFAR-10 and CIFAR-100 dataset for each ELU and PELU activation. The minimum median test error for ELU increases from 5.99\% and 25.08\% to 10.39\% and 34.75\% on CIFAR-10 and CIFAR-100 respectively, while for PELU it increases from 5.36\% and 24.55\% to 5.85\% and 25.38\%. We also observe that the relative error rate increase for our PELU is smaller than for ELU. Indeed, ELU has a relative minimum test error rate increase of 73\% and 39\% on CIFAR-10 and CIFAR-100 respectively, while PELU has 9\% and 3\%. Although this shows that our PELU parameterization reduces the detrimental effect of using BN before the activation, PELU should not be preceded by BN.

\subsection{ImageNet Object Recognition}
\label{ssec:experiment-imagenet}

\begin{figure*}[t]
\centering
\begin{minipage}{\linewidth}
\centering
\begin{minipage}{\linewidth}
\centering
\begin{minipage}{0.49\linewidth}
\centering
\includegraphics[width=\linewidth]{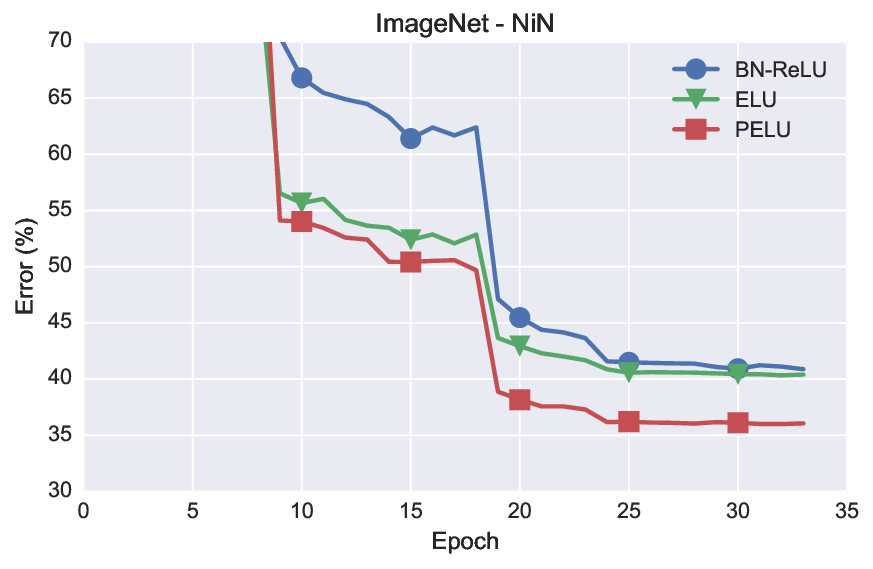}
\end{minipage}
\begin{minipage}{0.49\linewidth}
\centering
\includegraphics[width=\linewidth]{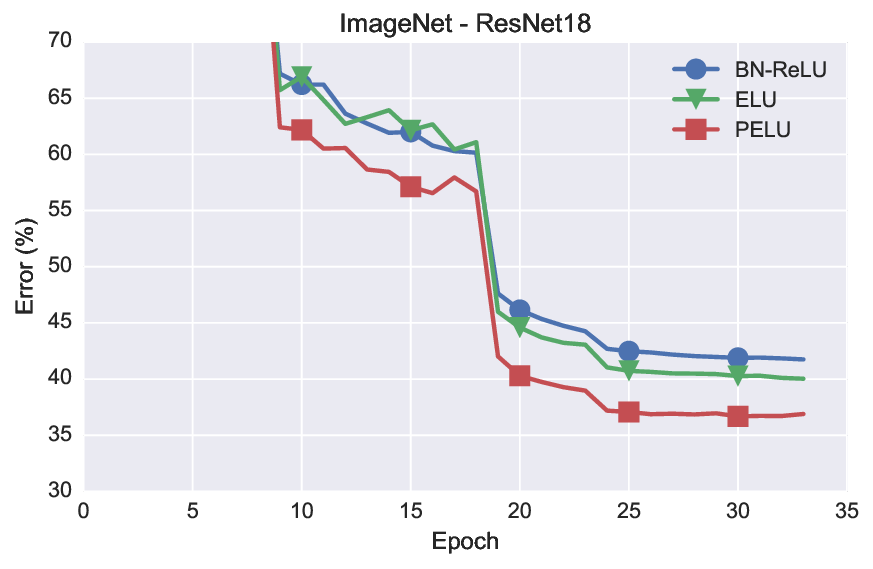} 
\end{minipage}
\end{minipage}
\begin{minipage}{\linewidth}
\centering
\begin{minipage}{0.49\linewidth}
\centering
\includegraphics[width=\linewidth]{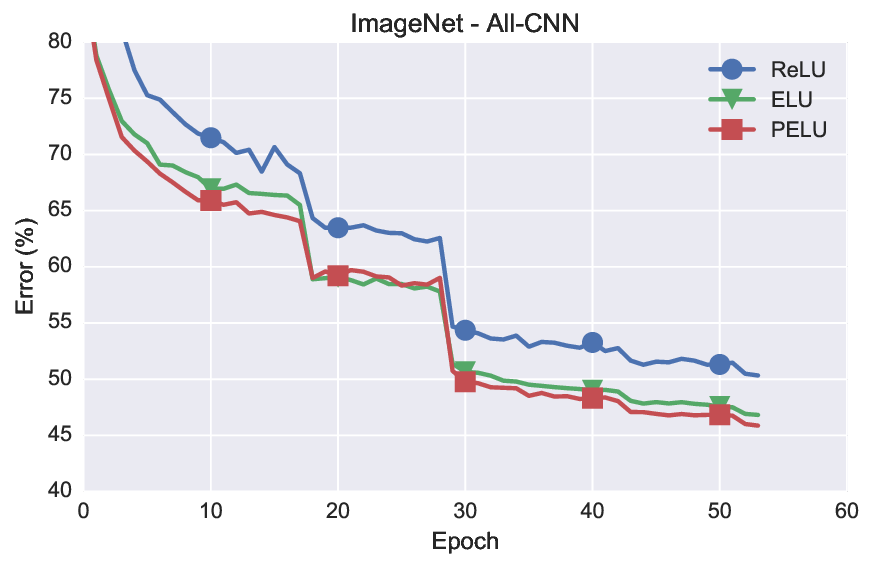}
\end{minipage}
\begin{minipage}{0.49\linewidth}
\centering
\includegraphics[width=\linewidth]{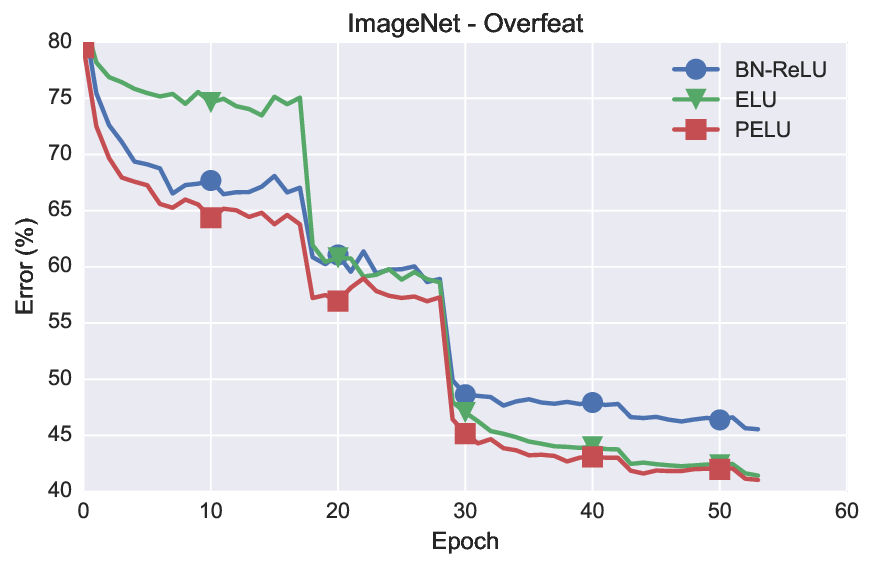}
\end{minipage}
\end{minipage}
\end{minipage}
\caption{TOP-1 error rate progression (in \%) of ResNet18, NiN, Overfeat and All-CNN on ImageNet 2012 validation set. NiN and ResNet18 (top row) used training regime \#1, while All-CNN and Overfeat (bottom row) used training regime \#2 (see Table~\ref{tab:imagenetTrainingRegime}). PELU has the lowest error rates for all networks. Regime \#1 shows a greater performance gap between ELU and PELU than regime \#2.}
\label{tab:imagenet-results}
\end{figure*}

We tested the proposed PELU on the ImageNet 2012 task (ILSVRC2012) using four different network architectures: ResNet18~\cite{shah2016deep}, Network in Network (NiN)~\cite{lin2013network}, All-CNN~\cite{springenberg2015striving} and Overfeat~\cite{sermanet2013overfeat}. The ResNet18 building block structure is shown in Figure~\ref{fig:resnet-structure}. In order not to favor PELU to the detriment of ELU and BN+ReLU, we performed minimal changes by only replacing the activation function. We used either PELU, ELU or BN+ReLU for the activation module. Each network was trained following Chintala's Torch implementation \verb|imagenet-multiGPU.torch| \footnote{ \url{https://github.com/soumith/imagenet-multiGPU.torch}} with the training regimes shown in Table~\ref{tab:imagenetTrainingRegime}. Regime \#1 starts at a higher learning rate than regime \#2, and has a larger learning rate decay.

\begin{table}[t]
\caption{ImageNet training regimes for modifying the learning rate and the weight decay. We trained NiN and ResNet18 using the first one, and trained All-CNN and Overfeat using the second one. The first regime starts at a higher learning rate, but has a larger decay than the second regime.}
\label{tab:imagenetTrainingRegime}
\centering
\begin{tabular}{ccp{3em}p{3em}p{3em}p{3em}}
\toprule
& & \multicolumn{4}{c}{Regime \#1 (ResNet18, NiN)} \\
\cmidrule{3-6} 
Epoch & & 1 & 10 & 20 & 25  \\
Learning Rate & & 1e-1 & 1e-2 & 1e-3 &  1e-4 \\
Weight Decay & & 5e-4 &  5e-4 &  0 &  0  \\
\bottomrule
\end{tabular}

\begin{tabular}{c}
{}
\end{tabular}

\begin{tabular}{ccp{3em}p{3em}p{3em}p{3em}c}
\toprule
& & \multicolumn{5}{c}{Regime \#2 (Overfeat, AllCNN)} \\
\cmidrule{3-6} 
Epoch & & 1 & 19 & 30 & 44 & 53 \\
Learning Rate & & 1e-2 & 5e-3 & 1e-3 & 5e-4 & 1e-4 \\
Weight Decay & & 5e-4 &  5e-4 & 0 &  0 &  0 \\
\bottomrule
\end{tabular}
\end{table}

Figure~\ref{tab:imagenet-results} presents the TOP-1 error rate (in \%) of all four networks on ImageNet 2012 validation dataset. In all cases, the networks using PELU outperformed the networks using ELU. NiN obtained the best result of 36.06\% with PELU, which corresponds to a relative improvement of 7.29\% compared to ELU (40.40\%). Since only 24 additional parameters were added to the network, this performance improvement indicates that PELU's parameterization acts in a different way than the weights and biases. Adding 24 additional weights throughout the network would not have been sufficient to increase the representative ability enough to get the observed performance improvement. Since such a low number of weights cannot significantly increase the expressive power of the network, these results indicate that the networks benefit from PELU.

%the fastest convergence rate, which is indicated by both the large plateaus between the learning rate epoch changes and the low error percentages. Even though Overfeat with PELU was only sightly better than with ELU, all networks with PELU outperformed those with ReLU and ELU. 

As shown in Figure~\ref{tab:imagenet-results}, the training regimes have an interesting effect on the convergence of the networks. The performance of PELU is closer to the performance of ELU for regime \#2, but is significantly better than ELU for regime \#1. We also observe that the error rates of All-CNN and Overfeat with PELU increase by a small amount starting at epoch 44, but stay steady for ELU and ReLU. These results suggest that training regimes with larger learning rates and decays help PELU to obtain a better performance improvement.

% It is worth noting that training Overfeat with ELU was difficult. Indeed, even though we changed the batch size to 64, 128 or 256, it did not converge and stayed at an error rate close to 99\%. This lack of convergence was caused by the presence of NaNs (not a number), indicating either overflows or division by 0. This was peculiar as we did not experience this problem with the other networks. On the other hand, we could see convergence when using BN before ELU or with small learning rate, but the network performed either poorly or converged slowly. Since Overfeat with PELU did not have this NaN problem, we concluded that a too negative ELU saturation was causing the problem. By using a saturation at $-0.5$ instead of $-1$, we managed to train the network without difficulty. Even though the NaN problem proved relatively easy to solve, this demonstrates the usefulness of adaptive activations like PELU that can deal automatically with these problems.

%Investigating the source of the error from the structure of Overfeat could be an interesting future work.

\subsection{Experimenting with Parameter Configuration}
\label{ssec:parameter-configuration}

\begin{figure}[t]
\centering
\begin{minipage}{\linewidth}
\includegraphics[width=\linewidth]{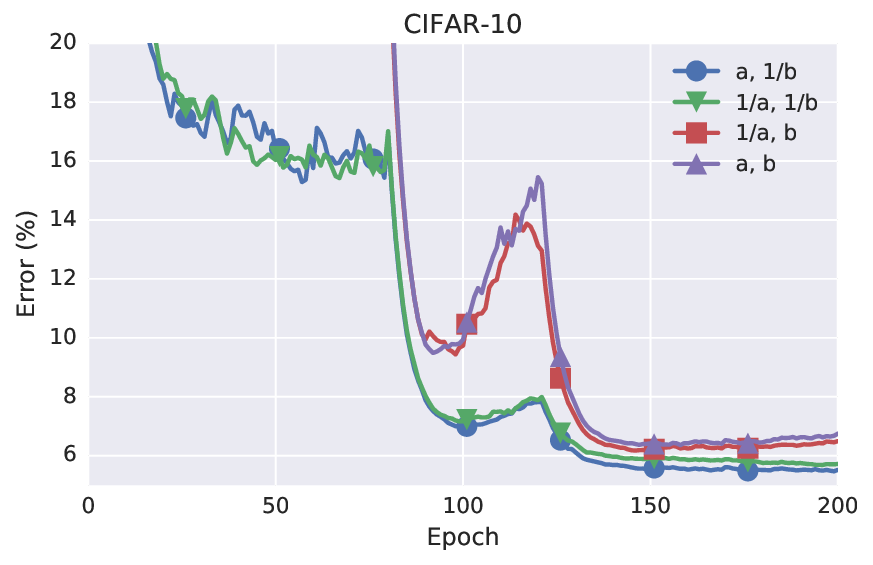}
\end{minipage}
\begin{minipage}{\linewidth}
\includegraphics[width=\linewidth]{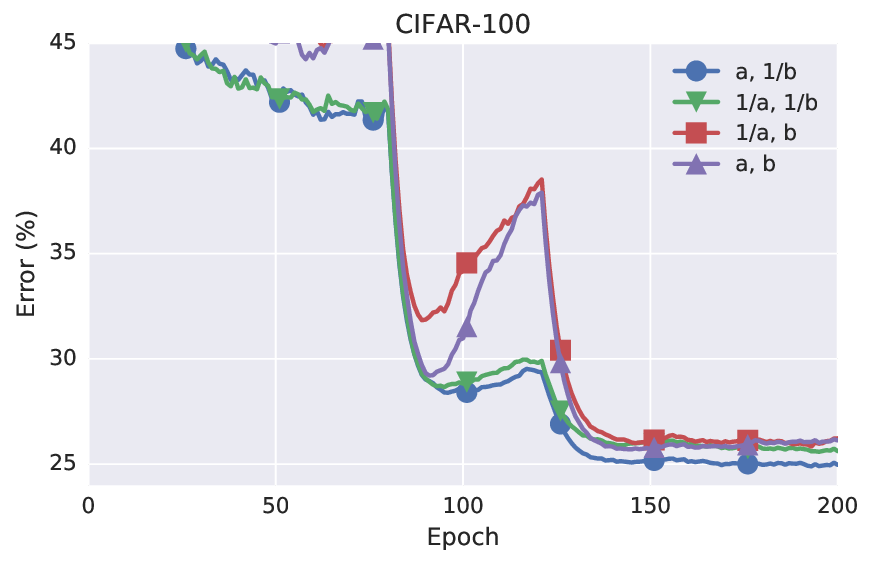}
\end{minipage}
\caption{Experimenting with the PELU parameter configuration in a ResNet with 110 layers on both CIFAR-10 and CIFAR-100 datasets. We show the convergence behavior of the median test error over five tries. Our proposed parameterization $(a, 1/b)$ obtained the best performance.}
\label{fig:pelu-div-mul-experiments}
\end{figure}

\begin{table}[t]
\centering
\caption{ResNet 110 layers test error (in \%) on both CIFAR-10 and CIFAR-100 datasets. We report the mean error over the last five epochs and the minimum error over all epochs (inside parenthesis) of the median error over five tries. We compare each four PELU configurations. Our proposed PELU configuration $(a, 1/b)$ obtained the best performance.}
\vspace{4pt}
\begin{tabular}{ccc}
\toprule
Configuration & CIFAR-10 & CIFAR-100 \\
\midrule
$(a,1/b)$ & \textbf{5.51 (5.36)} &  \textbf{25.02 (24.55)} \\ % div mul
$(1/a, 1/b)$ & 5.73 (5.60) & 25.68 (25.17) \\ % div div
$(1/a, b)$ & 6.51 (6.00) & 26.33 (25.48) \\ % mul div
$(a, b)$ & 6.74 (6.12) & 26.20 (25.24) \\ % mul mul
\bottomrule
\end{tabular}
\label{tab:pelu-div-mul-experiments}
\end{table}

The proposed PELU activation function~\eqref{eq:pelu} has two parameters $a$ and $b$, where $a$ is used with a multiplication and $b$ with a division. A priori, any of the four configurations $(a, b)$, $(a,1/b)$, $(1/a, b)$ or $(1/a, 1/b)$ could be used as parameterization. Note that these configurations are not reciprocal due to weight decay, which favors low weight magnitude. For instance, favoring low magnitude for parameter $b$ with the $(a, b)$ configuration favors a high PELU slope $a/b$. On the contrary, favoring low magnitude for parameter $b$ with the $(a,1/b)$ favors a low PELU slope $ab$. In order to better understand the difference between each configuration, we performed an experimental evaluation on the CIFAR-10 and CIFAR-100 datasets using the 110-layers ResNet as defined in Section~\ref{ssec:cifar-experiment}.

% call to get results: python show_results_v2.py

As shown in Table~\ref{tab:pelu-div-mul-experiments}, our proposed parameterization $(a, 1/b)$ obtained the best accuracy. Parameterization $(a, 1/b)$ obtained minimum test error medians of 5.36\% and 24.55\% on CIFAR-10 and CIFAR-100 respectively, while $(1/a, 1/b)$ obtained 5.60\% and 25.17\%, $(1/a, b)$ obtained 6.00\% and 25.48\%, and $(a,b)$ obtained 6.12\% and 25.24\%. These results also show that the two parameterizations with $1/b$ obtained a significantly lower error rate than the two parameterizations with $b$. From the convergence behavior in Figure~\ref{fig:pelu-div-mul-experiments}, we see that the parameterizations with $b$ have a larger error increase during the second stage of the training phase than parameterizations with $1/b$, and converge to lower error rates. These results concur with our observations  in Section~\ref{sec:pelu} of the effect of the parameters. Since weight decay pushes the weight magnitude towards zero, the function saturates slower as $b$ decreases with parameterizations using $b$. This encourages the function to have an almost linear shape on all its input values, which removes the non-linear characteristic of the activation. On the contrary, the function saturates faster a $b$ decreases with parameterizations using $1/b$. This helps the activation to keep its non-linear nature, which explains the observed performance gaps between $b$ and $1/b$.

\subsection{Parameter Progression}
\label{ssec:training-progression}

\begin{figure}[t!]
\centering
\begin{minipage}{\linewidth}
\centering
\begin{minipage}{0.49\linewidth}
\includegraphics[width=\linewidth]{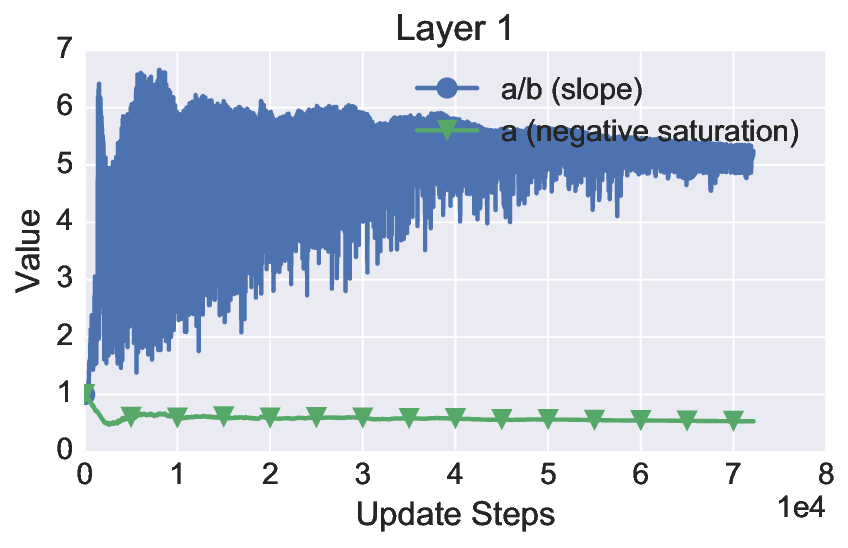}
\end{minipage}
\begin{minipage}{0.49\linewidth}
\includegraphics[width=\linewidth]{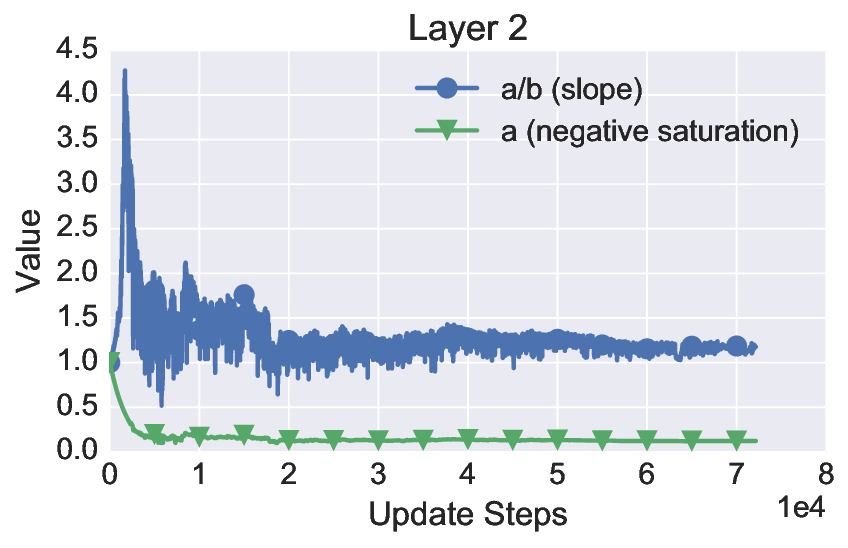}
\end{minipage}
\begin{minipage}{0.49\linewidth}
\includegraphics[width=\linewidth]{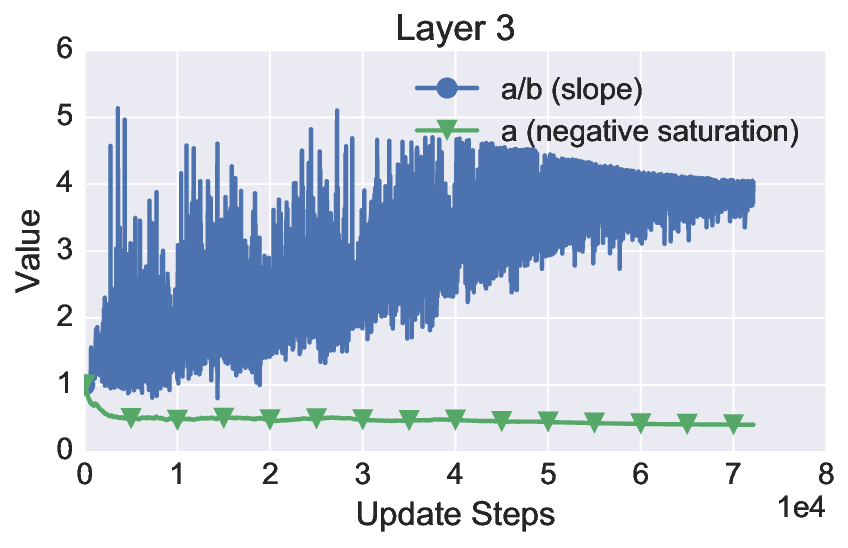}
\end{minipage}
\begin{minipage}{0.49\linewidth}
\includegraphics[width=\linewidth]{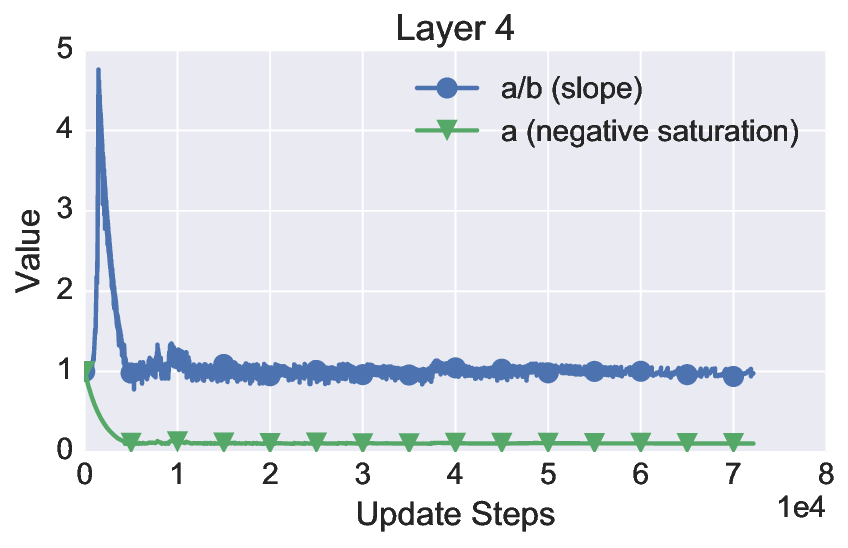}
\end{minipage}

\begin{minipage}{0.49\linewidth}
\includegraphics[width=\linewidth]{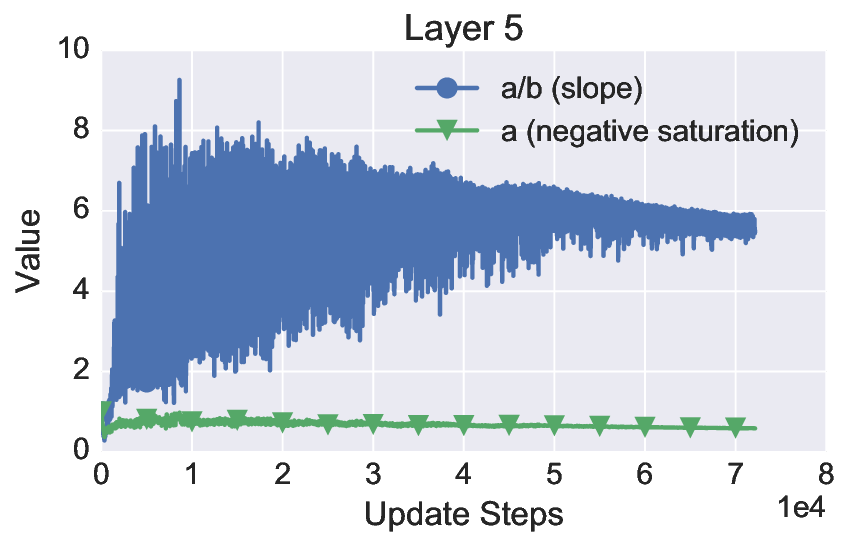}
\end{minipage}
\begin{minipage}{0.49\linewidth}
\includegraphics[width=\linewidth]{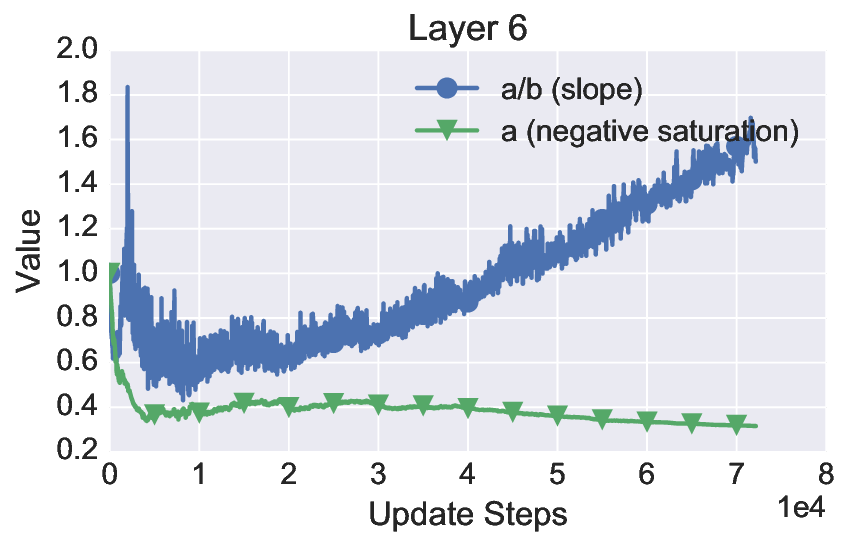}
\end{minipage}
\begin{minipage}{0.49\linewidth}
\includegraphics[width=\linewidth]{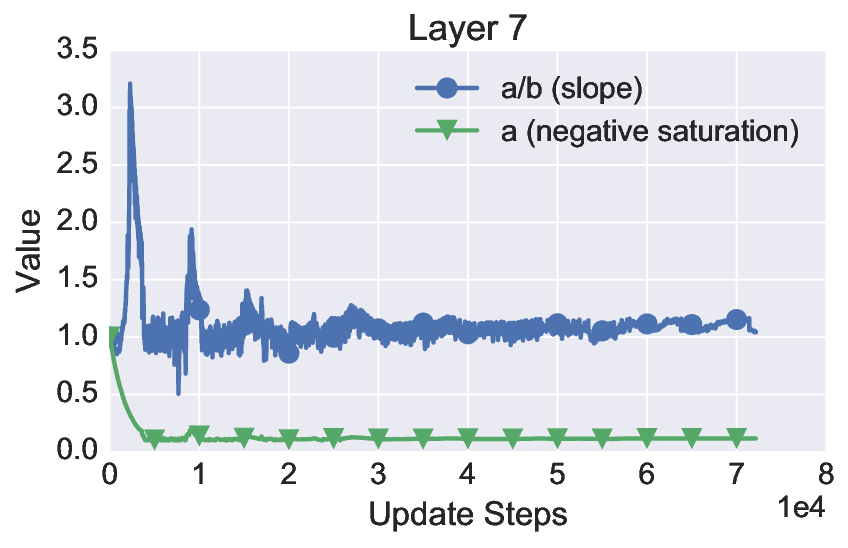}
\end{minipage}
\begin{minipage}{0.49\linewidth}
\includegraphics[width=\linewidth]{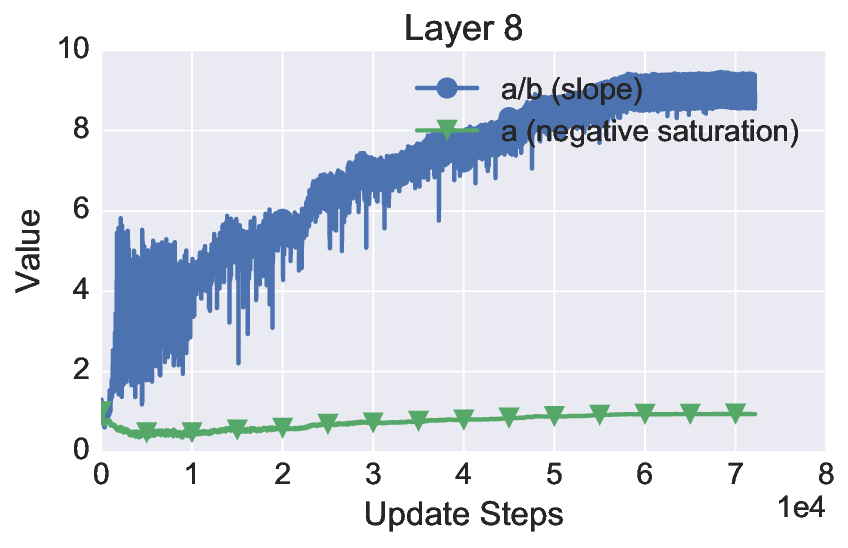}
\end{minipage}

\begin{minipage}{0.49\linewidth}
\includegraphics[width=\linewidth]{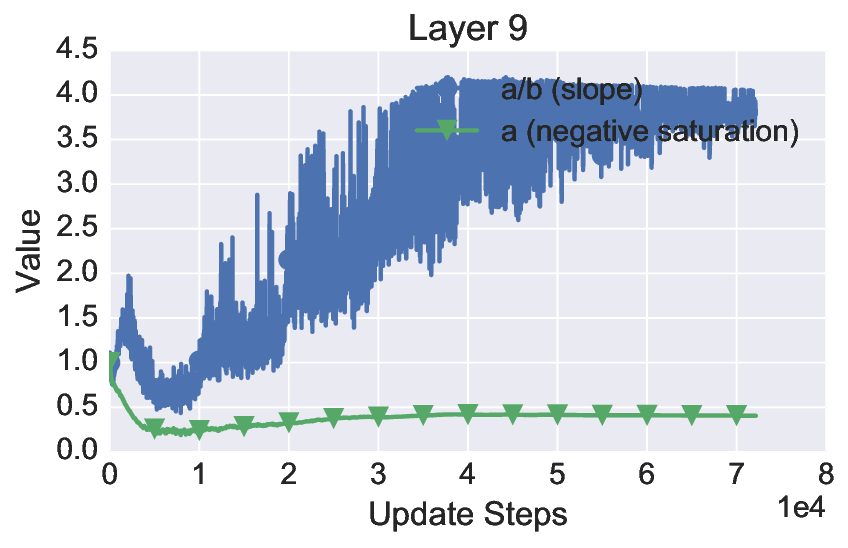}
\end{minipage}
\begin{minipage}{0.49\linewidth}
\includegraphics[width=\linewidth]{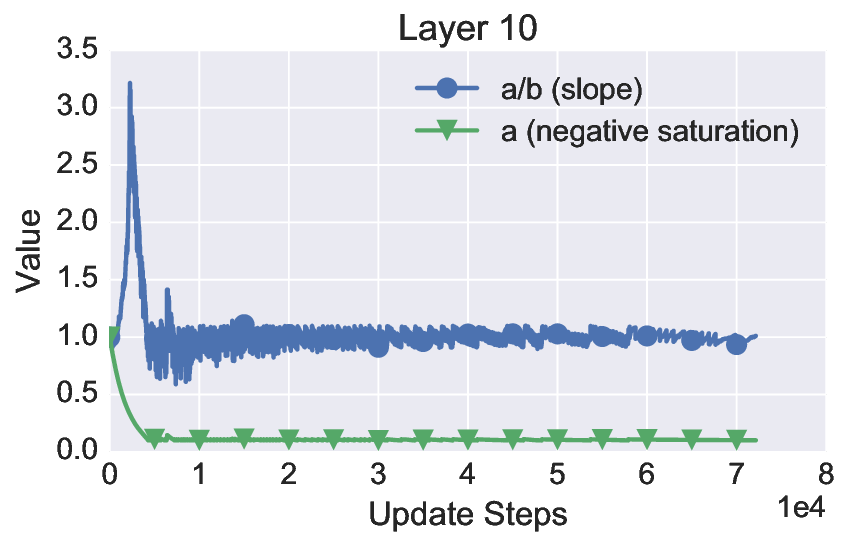}
\end{minipage}
\begin{minipage}{0.49\linewidth}
\includegraphics[width=\linewidth]{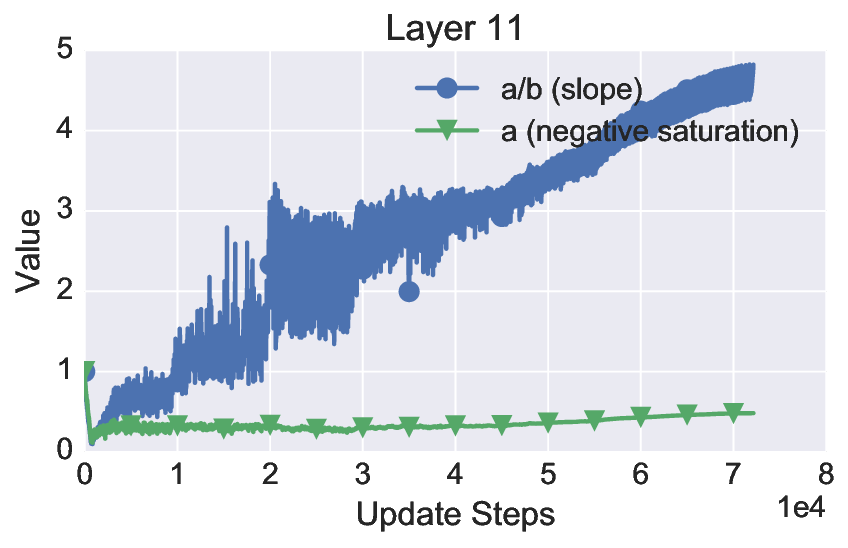}
\end{minipage}
\begin{minipage}{0.49\linewidth}
\includegraphics[width=\linewidth]{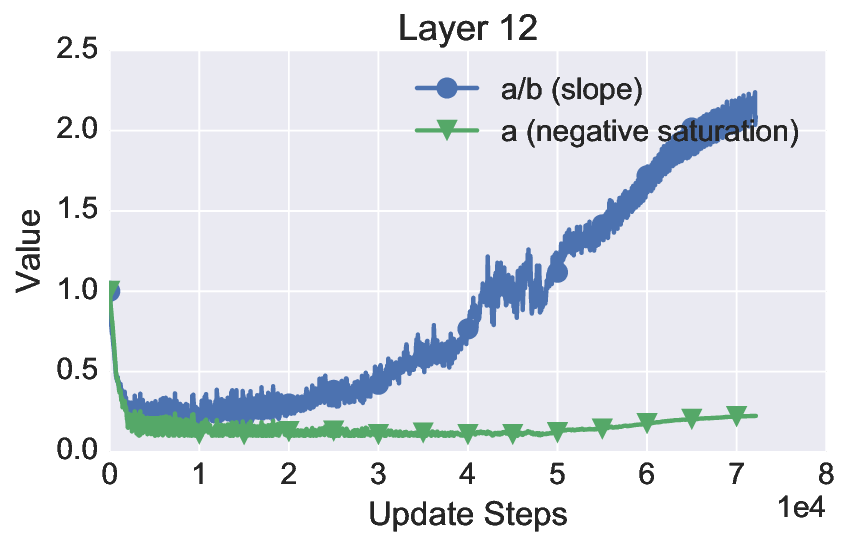}
\end{minipage}

\begin{minipage}{0.49\linewidth}
\includegraphics[width=\linewidth]{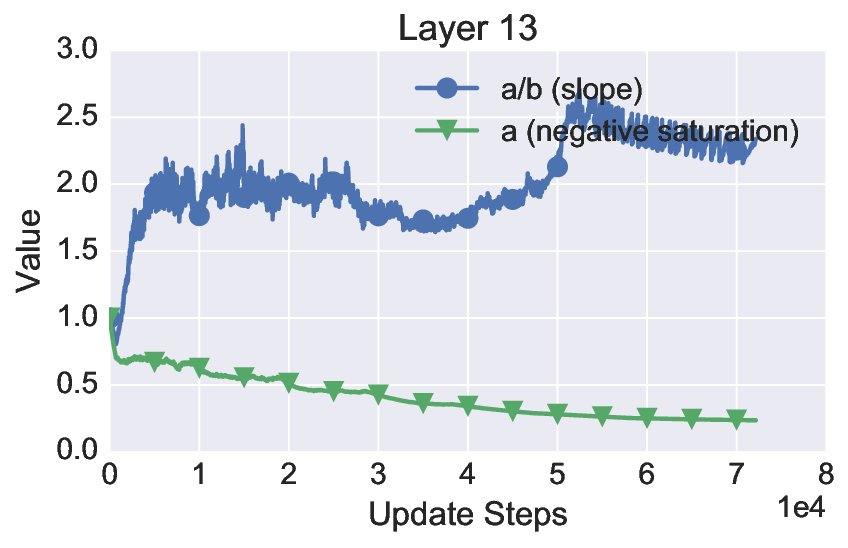}
\end{minipage}
\begin{minipage}{0.49\linewidth}
\includegraphics[width=\linewidth]{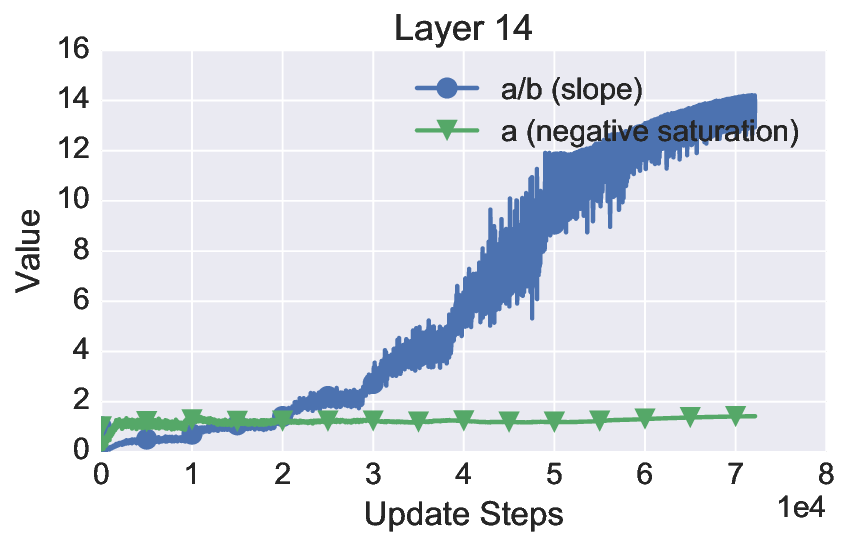}
\end{minipage}

\end{minipage}
\caption{PELU parameter progression at each layer of Vgg trained on CIFAR-10. We present the variation of the slope ($\frac{a}{b}$) and the negative saturation (parameter $a$). The network adopted different non-linear behaviors throughout the training phase.}
\label{fig:progressionParametersVggCifar10}
\end{figure}

We perform a visual evaluation of the non-linear behaviors adopted by a Vgg network during training on the CIFAR-10 dataset~\cite{simonyan2014very}. Figure~\ref{fig:progressionParametersVggCifar10} shows the progression of the slope ($\frac{a}{b}$) and the negative of the saturation point (parameter $a$) for PELU at each layer of Vgg. 

We can see different behaviors. At layers 2, 4, 7 and 10, the slope quickly increased to a large value, then decreased and converged at a value near 1. As for parameter a, it quickly converged to a value near 0. A slope near 1 and a negative saturation near 0 indicates that the network learned activations having the same shape as ReLU. This is an interesting result because ReLU has the important effect of promoting activation sparsity~\cite{nair2010rectified, maasrectifier}. Although we do not have a clear understanding to why the network increases the slope then decreases it before converging to ReLU, we believe that increasing the slope helps early during training to disentangle redundant neurons. Since peak activations scatter more the inputs than flat ones, spreading values may allow the network to declutter neurons activating similarly to the same input patterns.

Another interesting observation is that, apart from the \textit{ReLU} layers (layers 2, 4, 7 and 10), the negative saturations of all layers converged at values other than 0. For instance, parameter $a$ converges to a value near 0.5 at layer 1, while it converges to a value near 2 at layer 14. A negative saturation other than zero indicates that the learned PELU activations outputs negative values for negative arguments. The Vgg network had the possibility to learn all activation functions with a zero negative saturation (i.e. shaped like ReLU), but opted for a majority of activations with a non-zero negative saturation. Having activation functions with negative values has been previously analyzed in the context of the standard ELU activation, and it has been proposed that it helps to manage bias shift~\cite{clevert2015fast}. These results constitute an additional experimental evidence that this characteristic is important for the network.

\section{Discussion}
\label{sec:discussion}

During all our experiments with ELU and PELU, we did not use Batch Normalization (BN) before the activations. This is due to the detrimental effect of preceding PELU and ELU with BN, as we have observed in Section~\ref{ssec:effect-of-bn} with our ResNet experiments on CIFAR-10 and CIFAR-100. Although this detrimental effect has also been previously observed with ELU by Clevert and his coworkers~\cite{clevert2015fast}, it is unclear why BN before ELU and PELU increases error rate, but reduces error rate before ReLU. One important difference is that ReLU is positively scale invariant and ELU is not. Indeed, for ReLU we have $\max\{0, k x\} = k \max\{0, x\}$, where $k \geq 0$, while for ELU, which can be expressed as $\max\{0,x\} + \min\{0,\exp\{x\}-1\}$, we have $\min\{0,\exp\{k x\}-1\} \neq k \min\{0,\exp\{x\}-1\}$. The fact that ReLU is positively scale invariant and ELU is not may be part of the reason why BN before ReLU helps but harms before ELU. Given that BN performs mean and standard deviation scaling, followed by an affine transformation (scaled by $\gamma$ and shifted by $\beta$), using a positively scale invariant activation function may be essential for BN to properly reduce internal covariate shift~\cite{ioffe2015batch} or manage bias shift~\cite{clevert2015fast}. We could validate this hypothesis by experimenting with a new positively scale invariant activation function and observing whether BN helps or not. We leave this idea as future work.

\section{Conclusion}
\label{sec:conclusion}

% Sentence 1: Restate the topic
Object recognition is an essential ability for improving visual perception in automated vision systems performing complex scene understanding.
% Sentence 2: Restate thesis
In a recent work, the Exponential Linear Unit (ELU) has been proposed as a key element in Convolutional Neural Networks (CNNs) for reducing bias shift, but has the inconvenience of defining a parameter that must be set by hand.
% Other sentence: Summarize what you have proposed.
In this paper, we proposed the Parametric ELU (PELU) that alleviates this limitation by learning a parameterization of the ELU activation function. Our results on the CIFAR-10/100 and ImageNet datasets using the ResNet, NiN, All-CNN and Overfeat networks show that CNNs with PELU have better performance than CNNs with ELU. Our experiments with Vgg have shown that the network uses the added flexibility provided by PELU by learning different activation shapes at different locations in the network.
% Last sentence (if needed): Call to work
Parameterizing other activation functions, such as Softplus, Sigmoid or Tanh, could be worth investigating.

\section*{Acknowledgements}

We gratefully acknowledge the support of NVIDIA Corporation for providing the Tesla K20, K80 and Titan X for our experiments.

\bibliographystyle{IEEEtran}
\bibliography{bibliography}

\end{document}